\documentclass[runningheads,a4paper]{llncs}

\usepackage{amssymb}
\usepackage{graphicx}
\usepackage{authblk}
\usepackage{url}
\usepackage{amsmath}
\usepackage{algorithm}
\usepackage{bm}
\usepackage{wrapfig} 
\usepackage{lipsum} 
\usepackage[noend]{algpseudocode}
\makeatletter
\def\BState{\State\hskip-\ALG@thistlm}
\algnewcommand\algorithmicinput{\textbf{Input:}}
\algnewcommand\INPUT{\item[\algorithmicinput]}
\algnewcommand\algorithmicfulltrain{\hspace{3em} $\triangleright\triangleright\triangleright$\textbf{ Fully supervised training}}
\algnewcommand\FULLTRAIN{\item[\algorithmicfulltrain]}
\algnewcommand\algorithmicweaktrain{\hspace{3em} $\triangleright\triangleright\triangleright$\textbf{ Weakly supervised training}}
\algnewcommand\WEAKTRAIN{\item[\algorithmicweaktrain]}
\makeatother
\urldef{\mailsa}\path|{alfred.hofmann, ursula.barth, ingrid.haas, frank.holzwarth,|
\urldef{\mailsb}\path|anna.kramer, leonie.kunz, christine.reiss, nicole.sator,|
\urldef{\mailsc}\path|erika.siebert-cole, peter.strasser, lncs}@springer.com|    
\newcommand{\keywords}[1]{\par\addvspace\baselineskip
\noindent\keywordname\enspace\ignorespaces#1}
\begin{document}

\mainmatter  

\title{DeepEM: Deep 3D ConvNets With EM For Weakly Supervised Pulmonary Nodule Detection}

\titlerunning{DeepEM for Weakly Supervised Detection}

%
%
\author{Wentao Zhu$^{\star}$, Yeeleng S. Vang$^{\star}$, Yufang Huang$^{\dagger}$, and Xiaohui Xie$^{\star}$}

\authorrunning{Wentao Zhu, Yeeleng S. Vang, Yufang Huang, and Xiaohui Xie}
\institute{$^{\star}$University of California, Irvine \qquad $^{\dagger}$Lenovo AI Lab  \\
	\{wentaoz1,ysvang,xhx\}@uci.edu, yufang@lenovo.com}
%
%

\toctitle{Lecture Notes in Computer Science}
\tocauthor{Authors' Instructions}
\maketitle

\begin{abstract}
Recently deep learning has been witnessing widespread adoption in various medical image applications. However, training complex deep neural nets requires large-scale datasets labeled with ground truth, which are often unavailable in many medical image domains. For instance, to train a deep neural net to detect pulmonary nodules in lung computed tomography (CT) images, current practice is to manually label nodule locations and sizes in many CT images to construct a sufficiently large training dataset, which is costly and difficult to scale. On the other hand, electronic medical records (EMR) contain plenty of partial information on the content of each medical image. In this work, we explore how to tap this vast, but currently unexplored data source to improve pulmonary nodule detection. We propose DeepEM, a novel deep 3D ConvNet framework augmented with expectation-maximization (EM), to mine weakly supervised labels in EMRs for pulmonary nodule detection. Experimental results show that DeepEM can lead to 1.5\% and 3.9\% average improvement in free-response receiver operating characteristic (FROC) scores on LUNA16 and Tianchi datasets, respectively, demonstrating the utility of incomplete information in EMRs for improving deep learning algorithms.\footnote{https://github.com/uci-cbcl/DeepEM-for-Weakly-Supervised-Detection.git}
\end{abstract}

\section{Introduction}
Lung cancer is the most common cause of cancer-related death in men. Low-dose lung computed tomography (CT) screening provides an effective way for early diagnosis and can sharply reduce the lung cancer mortality rate. Advanced computer-aided diagnosis (CAD) systems are expected to have high sensitivities while maintaining low false positive rates to be truly useful. Recent advance in deep learning provides new opportunities to design more effective CAD systems to help facilitate doctors in their effort to catch lung cancer in their early stages.

The emergence of large-scale datasets such as the LUNA16 \cite{setio2017validation} has helped to accelerate research in nodule detection. Typically, nodule detection consists of two stages: nodule proposal generation and false positive reduction. Traditional approaches generally require hand-designed features such as morphological features, voxel clustering and pixel thresholding \cite{murphy2009large,jacobs2014automatic,lopez2015large}. More recently, deep convolutional architectures were employed to generate the candidate bounding boxes. Setio et al. proposed multi-view convolutional network for false positive nodule reduction \cite{setio2016pulmonarymultiview}. Several work employed 3D convolutional networks to handle the challenge due to the 3D nature of CT scans. The 3D fully convolutional network (FCN) was proposed to generate region candidates and deep convolutional network with weighted sampling was used in the false positive reduction stage \cite{zhu2018deeplung,dou2017automated,liao2017evaluate,tang18}. CASED proposed curriculum adaptive sampling for 3D U-net training in nodule detection \cite{jesson2017cased,ronneberger2015u}. Ding et al. used Faster R-CNN to generate candidate nodules, followed by 3D convolutional networks to remove false positive nodules \cite{ding2017accurate}. Due to the effective performance of Faster R-CNN \cite{kaimingfasterrcnn}, Faster R-CNN with a U-net-like encoder-decoder scheme was proposed for nodule detection \cite{zhu2018deeplung}. 
{\small{
\begin{figure}[t]
	\begin{center}
		\includegraphics[width=0.9\linewidth]{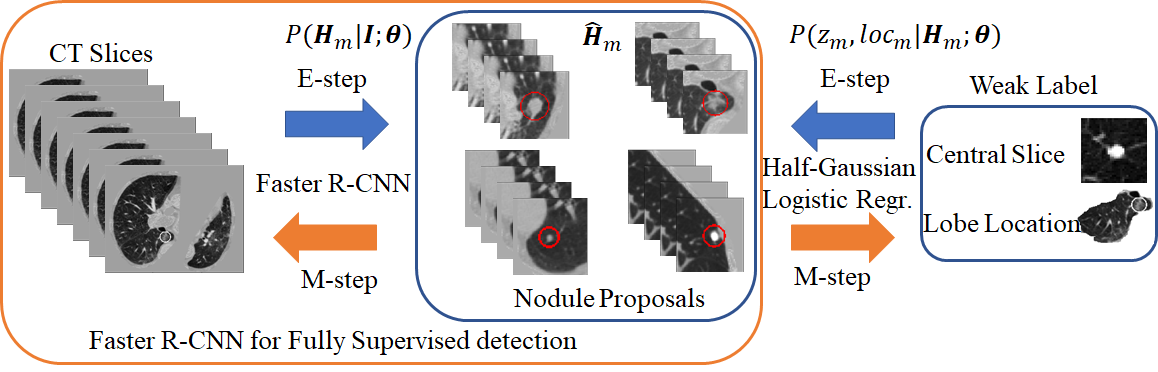}
		\caption{Illustration of DeepEM framework. Faster R-CNN is employed for nodule proposal generation. Half-Gaussian model and logistic regression are employed for central slice and lobe location respectively. In the E-step, we utilize all the observations, CT slices, and weak label to infer the latent variable, nodule proposals, by maximum a posteriori (MAP) or sampling. In the M-step, we employ the estimated proposals to update parameters in the Faster R-CNN and logistic regression.}
		\label{fig:framework}
	\end{center}
\end{figure}
}}

A prerequisite to utilization of deep learning models is the existence of an abundance of labeled data. However, labels are especially difficult to obtain in the medical image analysis domain. There are multiple contributing factors: a) labeling medical data typically requires specially trained doctors; b) marking lesion boundaries can be hard even for experts because of  low signal-to-noise ratio in many medical images; and c) for CT and magnetic resonance imaging (MRI) images, the annotators need to label the entire 3D volumetric data, which can be costly and time-consuming. Due to these limitations, CT medical image datasets are usually small, which can lead to over-fitting on the training set and, by extension, poor generalization performance on test sets \cite{zhu2018adversarial}.

By contrast, medical institutions have large amount of weakly labeled medical images. In these databases, each medical image is typically associated with an electronic medical report (EMR). Although these reports may not contain explicit information on detection bounding box or segmentation ground truth, it often includes the results of diagnosis, rough locations and summary descriptions of lesions if they exist. We hypothesize that these extra sources of weakly labeled data may be used to enhance the performance of existing detector and improve its generalization capability. 


There are previous attempts to utilize weakly supervised labels to help train machine learning models. Deep multi-instance learning was proposed for lesion localization and whole mammogram classification \cite{zhu2017deep}. The two-stream spatio-temporal ConvNet was proposed to recognize heart frames and localize the heart using only weak labels for whole ultrasound image of fetal heartbeat \cite{gao2017detection}. Different pooling strategies were proposed for weakly supervised localization and segmentation respectively \cite{wang2017chestx,feng2017discriminative,bilen2016weakly}. Papandreou et al. proposed an iterative approach to infer pixel-wise label using image classification label for segmentation \cite{papandreou2015weakly}. Self-transfer learning co-optimized both classification and localization networks for weakly supervised lesion localization \cite{hwang2016self}. Different from these works, we consider nodule proposal as latent variable and propose DeepEM, a new deep 3D convolutional nets with Expectation-Maximization optimization, to mine the big data source of weakly supervised label in EMR as illustrated in Fig. \ref{fig:framework}. Specifically, we infer the posterior probabilities of the proposed nodules being true nodules, and utilize the posterior probabilities to train nodule detection models. 

\section{DeepEM for Weakly Supervised Detection}
\textbf{Notation} We denote by $\bm{I} \in \mathbb{R}^{h \times w \times s}$ the CT image, where $h$, $w$, and $s$ are image height, width, and number of slices respectively. The nodule bounding boxes for $\bm{I}$ are denoted as ${\bm{H}}=\{\bm{H}_1, \bm{H}_2, \dots, \bm{H}_M\}$, where $\bm{H}_m = \{x_m, y_m, z_m, d_m\}$, the $(x_m, y_m, z_m)$ represents the center of nodule proposal, $d_m$ is the diameter of the nodule proposal, and $M$ is the number of nodules in the image $\bm{I}$. In the weakly supervised scenario, the nodule proposal $\bm{H}$ is a latent variable, and each image $\bm{I}$ is associated with weak label ${\bm{X}}=\{\bm{X}_1, \bm{X}_2, \dots, \bm{X}_M\}$, where $\bm{X}_m=\{{loc}_m, z_m\}$, ${loc}_m \in \{1,2,3,4,5,6\}$ is the location (right upper lobe, right middle lobe, right lower lobe, left upper lobe, lingula, left lower lobe) of nodule $\bm{H}_m$ in the lung, and $z_m$ is the central slice of the nodule. 

For fully supervised detection, the objective function is to maximize the log-likelihood function for observed nodule ground truth $\bm{H}$ given image $\bm{I}$ as 
\begin{equation}
	\mathcal{L}(\bm{\theta}) = \log P(\bm{H} \cup \bm{\bar{H}} | \bm{I}; \bm{\theta}) = \frac{1}{M}\sum_{m=1}^{M} \log P(\bm{H}_m | \bm{I}; \bm{\theta})+ \frac{1}{N}\sum_{n=1}^{N} \log P(\bm{\bar{H}}_n | \bm{I}; \bm{\theta}) ,\label{eq:loglike} %
\end{equation}
where $\bm{\bar{H}}=\{\bm{\bar{H}}_1, \bm{\bar{H}}_2, \dots, \bm{\bar{H}}_N\}$ are hard negative nodule proposals \cite{kaimingfasterrcnn}, $\bm{\theta}$ is the weights of deep 3D ConvNet. We employ Faster R-CNN with 3D Res18 for the fully supervised detection because of its superior performance.

For weakly supervised detection, nodule proposal $\bm{H}$ can be considered as a latent variable. Using this framework, image $\bm{I}$ and weak label \(\bm{X}=\{({loc}_1, z_1), ({loc}_2, \\ z_2), \dots, ({loc}_M, z_M)\}\) can be considered as observations. The joint distribution is
\begin{equation}
\begin{aligned}
P(\bm{I}, \bm{H}, \bm{X}; \bm{\theta}) &= P(\bm{I}) \prod_{m=1}^{M} \big( P(\bm{H}_m|\bm{I}; \bm{\theta}) P(\bm{X}_m | \bm{H}_m; \bm{\theta}) \big) \\
&= P(\bm{I}) \prod_{m=1}^{M} \big( P(\bm{H}_m|\bm{I}; \bm{\theta}) P({loc}_m | \bm{H}_m; \bm{\theta}) P({z}_m | \bm{H}_m; \bm{\theta}) \big).\label{eq:pgm} 
\end{aligned}
\end{equation}
To model $P({z}_m | \bm{H}_m; \bm{\theta})$, we propose using a half-Gaussian distribution based on nodule size distribution because $z_m$ is correct if it is within the nodule area (center slice of $\bm{H}_m$ as ${z}_{{\bm{H}}_m}$, and nodule size $\sigma$ can be empirically estimated based on existing data) for nodule detection in Fig. \ref{fig:halfnorm}(a). For lung lobe prediction $P({loc}_m | \bm{H}_m; \bm{\theta})$, a logistic regression model is used based on relative value of nodule center $({x}_{{\bm{H}}_m}, {y}_{{\bm{H}}_m}, {z}_{{\bm{H}}_m})$ after lung segmentation. That is
\begin{equation}
	\begin{aligned}
	P(z_m, {loc}_m | \bm{H}_m ; \bm{\theta}) = \frac{2}{\sqrt{2 \pi {\sigma}^2}} \exp \big( -\frac{|z_m - {z}_{\bm{H}_m} |^2}{2 {\sigma}^2} \big) \frac{\exp(\bm{f}(\bm{H}_m) \bm{\theta}_{{loc}_m})}{\sum_{{{loc}_m}=1}^{6}\exp(\bm{f}(\bm{H}_m) \bm{\theta}_{{loc}_m})},\label{eq:weaklikeli}
	\end{aligned}
\end{equation}
where $\bm{\theta}_{{loc}_m}$ is the associated weights with lobe location ${loc}_m$ for logistic regression, feature $\bm{f}(\bm{H}_m) = (\frac{{x}_{{\bm{H}}_m}}{{x}_{\bm{I}}}, \frac{{y}_{{\bm{H}}_m}}{{y}_{\bm{I}}}, \frac{{z}_{{\bm{H}}_m}}{{z}_{\bm{I}}})$, and $({x}_{\bm{I}}, {y}_{\bm{I}}, {z}_{\bm{I}})$ is the total size of image $\bm{I}$ after lung segmentation. In the experiments, we found the logistic regression converges quickly and is stable.

The expectation-maximization (EM) is a commonly used approach to optimize the maximum log-likelihood function when there are latent variables in the model. We employ the EM algorithm to optimize deep weakly supervised detection model in equation \ref{eq:pgm}. The expected complete-data log-likelihood function given previous estimated parameter ${\bm{\theta}}^{\prime}$ in deep 3D Faster R-CNN is
\begin{equation}
\begin{aligned}
Q(\bm{\theta}; \bm{\theta^{\prime}}) = & \frac{1}{M}\sum_{m=1}^{M}  \mathbb{E}_{P(\bm{H}_m | \bm{I}, z_m, {loc}_m; {\bm{\theta}}^{\prime})} \big[ \log P(\bm{H}_m|\bm{I}; \bm{\theta}) \\ &+ \log P(z_m, {loc}_m|{\bm{H}}_m; \bm{\theta}) \big] +  \mathbb{E}_{Q(\bm{\bar{H}}_n | \bm{z})}\big[ \log P(\bm{\bar{H}}_n | \bm{I}; \bm{\theta}) \big], \label{eq:expect}
\end{aligned}
\end{equation}
where $\bm{z} = \{z_1, z_2, \dots, z_m\}$. In the implementation, we only keep hard negative proposals far away from weak annotation $\bm{z}$ to simplify $Q(\bm{\bar{H}}_n | \bm{z})$. The posterior distribution of latent variable $\bm{H}_m$ can be calculated by
\begin{equation}
\begin{aligned}
P(\bm{H}_m | \bm{I}, z_m, {loc}_m; \bm{{\theta}^{\prime}}) &\propto P(\bm{H}_m | \bm{I}; \bm{{\theta}^{\prime}}) P(z_m, {loc}_m | \bm{H}_m; \bm{{\theta}^{\prime}}).\label{eq:posteri}
\end{aligned}
\end{equation}
Because Faster R-CNN yields a large number of proposals, we first use hard threshold (-3 before sigmoid function) to remove proposals of small confident probability, then employ non-maximum suppression (NMS) with intersection over union (IoU) as 0.1. We then employ two schemes to approximately infer the latent variable $\bm{H}_m$: maximum a posteriori (MAP) or sampling. \\ 
\textbf{DeepEM with MAP} We only use the proposal of maximal posterior probability to calculate the expectation.
\begin{equation}
	\hat{\bm{H}}_m = {\arg \max }_{\bm{H}_m} P(\bm{H}_m | \bm{I}; \bm{{\theta}^{\prime}}) P(z_m, {loc}_m | \bm{H}_m; \bm{{\theta}^{\prime}})\label{eq:map} 
\end{equation} 
\textbf{DeepEM with Sampling} We approximate the distribution by sampling $\hat{M}$ proposals $\hat{\bm{H}}_m$ according to normalized equation \ref{eq:posteri}. The expected log-likelihood function in equation \ref{eq:expect} becomes
\begin{equation}
\begin{aligned}
Q(\bm{\theta}; \bm{\theta^{\prime}}) = &\frac{1}{M \hat{M}}\sum_{m=1}^{M}  \sum_{\hat{\bm{H}}_m}^{\hat{M}} \big( \log P(\hat{\bm{H}}_m|\bm{I}; \bm{\theta}) + \log P(z_m, {loc}_m|{\hat{\bm{H}}}_m; \bm{\theta}) \big) \\ & + \mathbb{E}_{Q(\bm{\bar{H}}_n | \bm{z})}\big[ \log P(\bm{\bar{H}}_n | \bm{I}; \bm{\theta}) \big]. \label{eq:deepemsamp}
\end{aligned}
\end{equation}

After obtaining the expectation of complete-data log-likelihood function in equation \ref{eq:expect}, we can update the parameters $\bm{\theta}$ by 
\begin{equation}
	\hat{\bm{\theta}} = \arg \max Q(\bm{\theta} ; {\bm{\theta}}^{\prime}).\label{eq:mstep}
\end{equation}
The M-step in equation \ref{eq:mstep} can be conducted by stochastic gradient descent commonly used in deep network optimization for equation \ref{eq:loglike}. Our entire algorithm is outlined in algorithm \ref{alg:deepem}.
{\small{
\begin{algorithm}[t]
	\caption{DeepEM for Weakly Supervised Detection}\label{alg:deepem}
	\begin{algorithmic}[1]
		\INPUT Fully supervised dataset $D_F = \{({\bm{I}}, {\bm{H}})_i\}_{i=1}^{N_F}$, weakly supervised dataset $D_W = \{({\bm{I}}, {\bm{X}})_i\}_{i=1}^{N_W}$, 3D Faster R-CNN and logistic regression parameters $\bm{\theta}$.
		\BState \emph{Initialization}: Update weights $\bm{\theta}$ by maximizing equation \ref{eq:loglike} using data from $D_F$.
		\BState \emph{for epoch = 1 to \#TotalEpochs}:
		\WEAKTRAIN
		\State \hspace{1.3em} Use Faster R-CNN model ${\bm{\theta}}^{\prime}$ to obtain proposal probability $P(\bm{H}_m | \bm{I}; \bm{{\theta}^{\prime}})$ for weakly supervised data sampled from $D_W$.
		\State \hspace{1.3em} Remove proposals with small probabilities and NMS. 
		\BState \hspace{1.3em} \emph{for m = 1 to M}:   \hspace{1.3em} $\triangleright\triangleright\triangleright$ Each weak label
		\State \hspace{3.0em} Calculate $P(z_m, {loc}_m | \bm{H}_m ; \bm{\theta})$ for each proposal by equation \ref{eq:weaklikeli}.
		\State \hspace{3.0em} Estimate posterior distribution $P(\bm{H}_m | \bm{I}, z_m, {loc}_m; \bm{{\theta}^{\prime}})$ by equation \ref{eq:posteri} with normalization.
		\State \hspace{3.0em} Employ MAP by equation \ref{eq:map} or Sampling to obtain the inference of $\bm{H}_m$.
		\State \hspace{1.3em} Obtain the expect log-likelihood function by equation \ref{eq:expect} using the estimated proposal (MAP) or by equation \ref{eq:deepemsamp} (Sampling).
		\State \hspace{1.3em} Update parameter by equation \ref{eq:mstep}.
		\FULLTRAIN
		\State	\hspace{1.3em} Update weights $\bm{\theta}$ by maximizing equation \ref{eq:loglike} using fully supervised data $D_F$. 
	\end{algorithmic}
\end{algorithm}
}}
\section{Experiments}
We used 3 datasets, LUNA16 dataset \cite{setio2017validation} as fully supervised nodule detection, NCI NLST\footnote{https://biometry.nci.nih.gov/cdas/datasets/nlst/} dataset as weakly supervised detection, Tianchi Lung Nodule Detection\footnote{https://tianchi.aliyun.com/} dataset as holdout dataset for test only. LUNA16 dataset is the largest publicly available dataset for pulmonary nodules detection \cite{setio2017validation}. LUNA16 dataset removes CTs with slice thickness greater than 3mm, slice spacing inconsistent or missing slices, and consist of 888 low-dose lung CTs which have explicit patient-level 10-fold cross validation split. NLST dataset consists of hundreds of thousands of lung CT images associated with electronic medical records (EMR). In this work, we focus on nodule detection based on image modality and only use the central slice and nodule location as weak supervision from the EMR. As part of data cleansing, we remove negative CTs, CTs with slice thickness greater than 3mm and nodule diameter less than 3mm. After data cleaning, we have 17,602 CTs left with 30,951 weak annotations. In each epoch, we randomly sample $\frac{1}{16}$ CT images for weakly supervised training because of the large numbers of weakly supervised CTs. Tianchi dataset contains 600 training low-dose lung CTs and 200 validation low-dose lung CTs for nodule detection. The annotations are location centroids and diameters of the pulmonary nodules, and do not have less than 3mm diameter nodule, which are the same with those on LUNA16 dataset. 

\textbf{Parameter estimation in $P({z}_m | \bm{H}_m; \bm{\theta})$} 
If the current $z_m$ is within the nodule, it is a true positive proposal. We can model $|z_m-z_{{\bm{H}}_m}|$ using a half-Gaussian distribution shown as the red dash line in Fig. \ref{fig:halfnorm}(a). The parameters of the half-Gaussian is estimated from the LUNA16 data empirically. Because LUNA16 removes nodules of diameter less than 3mm, we use the truncated half-Gaussian to model the central slice $z_m$ as $\max(|z_m-z_{{\bm{H}}_m}|-\mu, 0)$, where $\mu$ is the mean of related Gaussian as the minimal nodule radius with 1.63. 
{\small{
\begin{figure}[t]
	\begin{center}
		\begin{minipage}{0.36\textwidth}
			\centerline{\includegraphics[width=0.9\linewidth]{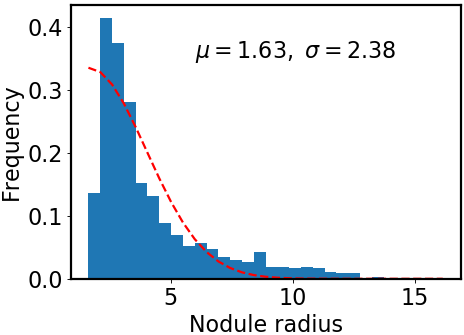}}
			\center{(a)}
		\end{minipage}
		\begin{minipage}{0.63\textwidth}
			\centerline{\includegraphics[width=1.08\linewidth]{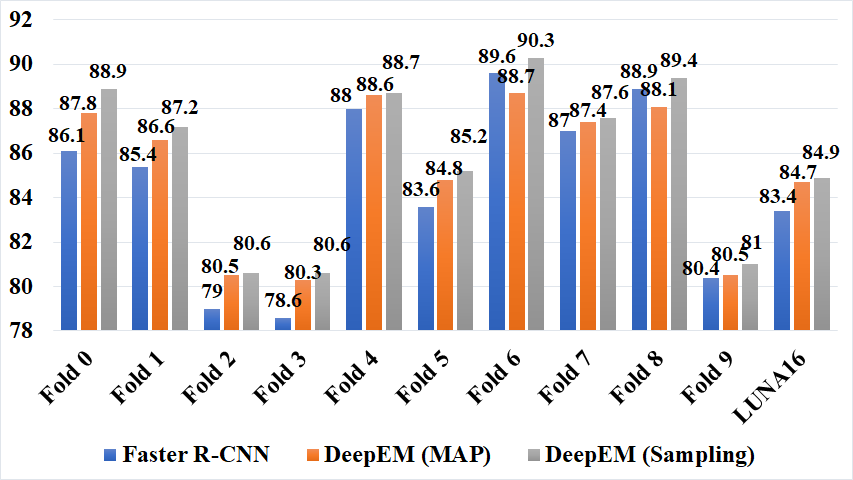}}
			\center{(b)}
		\end{minipage}
		\caption{(a)Empirical estimation of half-Gaussian model for $P({z}_m | \bm{H}_m; \bm{\theta})$ on LUNA16. (b) FROC (\%) comparison among Faster R-CNN, DeepEM with MAP, DeepEM with Sampling on LUNA16.}
		\label{fig:halfnorm}
	\end{center}
\end{figure}}}

\textbf{Performance comparisons on LUNA16} 
We conduct 10-fold cross validation on LUNA16 to validate the effectiveness of DeepEM. The baseline method is Faster R-CNN with 3D Res18 network denoted as \textbf{Faster R-CNN} \cite{kaimingfasterrcnn,zhu2018deeplung}. Then we employ it to model $P(\bm{H}_m | \bm{I}; \bm{{\theta}^{\prime}})$ for weakly supervised detection scenario. Two inference scheme for ${\bm{H}}_m$ are used in DeepEM denoted as \textbf{DeepEM (MAP)} and \textbf{DeepEM (Sampling)}. In the proposal inference of DeepEM with Sampling, we sample two proposals for each weak label because the average number of nodules each CT is 1.78 on LUNA16. The evaluation metric, Free receiver operating characteristic (FROC), is the average recall rate at the average number of false positives at 0.125, 0.25, 0.5, 1, 2, 4, 8 per scan, which is the official evaluation metric for LUNA16 and Tianchi \cite{setio2017validation}. 

From Fig. \ref{fig:halfnorm}(b), DeepEM with MAP improves about 1.3\% FROC over Faster R-CNN and DeepEM with Sampling improves about 1.5\% FROC over Faster R-CNN on average on LUNA16 when incorporating weakly labeled data from NLST. We hypothesize the greater improvement of DeepEM with Sampling over DeepEM with MAP is that MAP inference is greedy and can get stuck at a local minimum while the nature of sampling may allow DeepEM with Sampling to escape these local minimums during optimization.

\textbf{Performance comparisons on holdout test set from Tianchi}
We employed a holdout test set from Tianchi to validate each model from 10-fold cross validation on LUNA16. The results are summarized in Table \ref{tab:tianchifrocperformance}. We can see DeepEM utilizing weakly supervised data improves 3.9\% FROC on average over Faster R-CNN. The improvement on holdout test data validates DeepEM as an effective model to exploit potentially large amount of weak data from electronic medical records (EMR) which would not require further costly annotation by expert doctors and can be easily obtained from hospital associations.
{\small{
\begin{table}[t]
	\centering
	\caption{FROC (\%) comparisons among Faster R-CNN with 3D ResNet18, DeepEM with MAP, DeepEM with Sampling on Tianchi.}\label{tab:tianchifrocperformance}
	\begin{tabular}{c|c|c|c|c|c|c|c|c|c|c|c}
		\hline
		Fold&0&1&2&3&4&5&6&7&8&9&Average\\
		\hline
		Faster R-CNN&72.8&70.8&69.8&71.9&76.4&73.0&71.3&74.7&72.9&71.3&72.5\\
		\hline
		DeepEM (MAP)&77.2&75.8&75.8&74.9&77.0&75.5&77.2&75.8&76.0&74.7&76.0\\
		\hline
		DeepEM (Sampling)&77.4&75.8&75.9&75.0&77.3&75.0&77.3&76.8&77.7&75.8&76.4\\
		\hline
	\end{tabular}
\end{table}}}
{\small{
\begin{figure}[t]
	\begin{center}
		\centerline{\includegraphics[width=\linewidth]{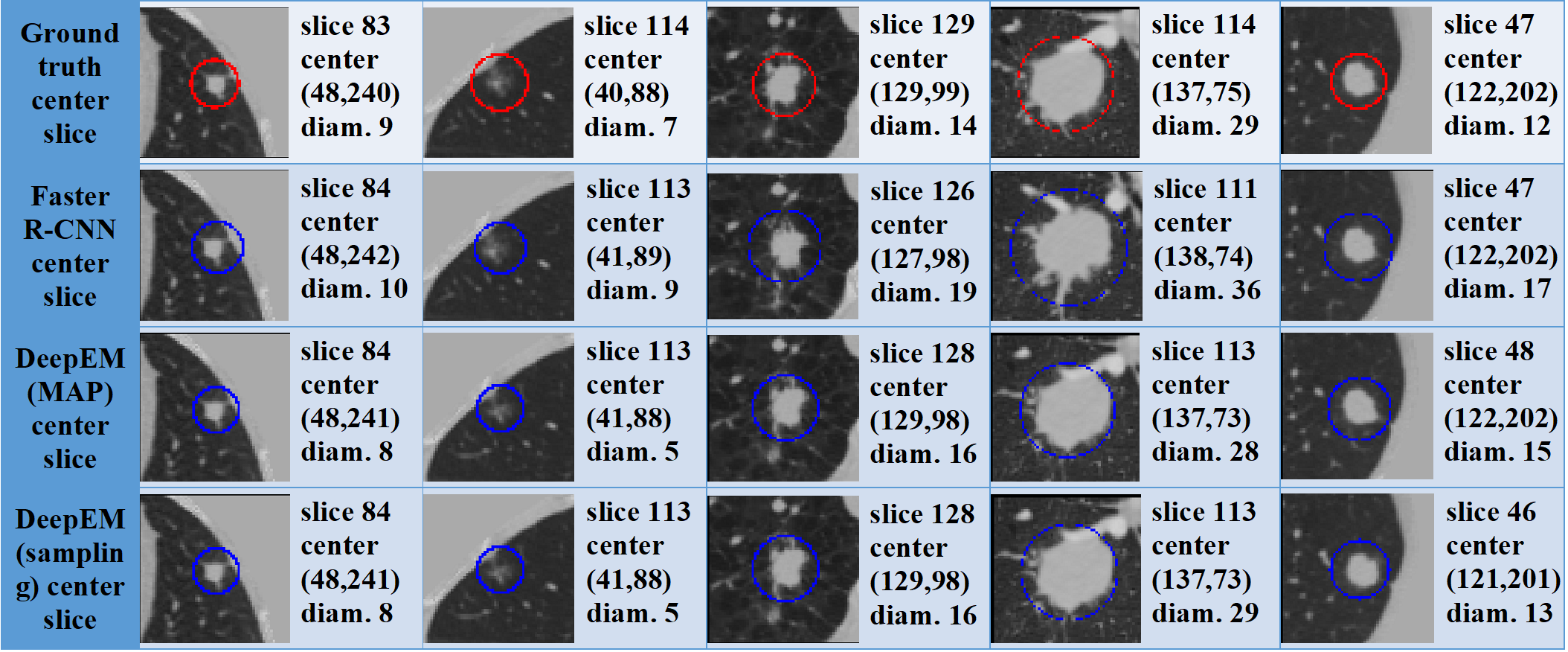}}
		\caption{Detection visual comparison among Faster R-CNN, DeepEM with MAP and DeepEM with Sampling on nodules randomly sampled from Tianchi. DeepEM provides more accurate detection (central slice, center and diameter) than Faster R-CNN.}
		\label{fig:vis}
	\end{center}
\end{figure}}}

\textbf{Visualizations} 
We compare Faster R-CNN with the proposed DeepEM visually in Fig. \ref{fig:halfnorm}(b). We randomly choose nodules from Tianchi. From Fig. \ref{fig:halfnorm}(b), DeepEM yields better detection for nodule center and tighter nodule diameter which demonstrates DeepEM improves the existing detector by exploiting weakly supervised data.
\section{Conclusion}
In this paper, we have focused on the problem of detecting pulmonary nodules from lung CT images, which previously has been formulated as a supervised learning problem and requires a large amount of training data with the locations and sizes of nodules precisely labeled. Here we propose a new framework, called DeepEM, for pulmonary nodule detection by taking advantage of abundantly available weakly labeled data extracted from EMRs. We treat each nodule proposal as a latent variable, and infer the posterior probabilities of proposal nodules being true ones conditioned on images and weak labels. The posterior probabilities are further fed to the nodule detection module for training. We use an EM algorithm to train the entire model end-to-end. Two schemes, maximum a posteriori (MAP) and sampling, are used for the inference of proposals. Extensive experimental results demonstrate the effectiveness of DeepEM for improving current state of the art nodule detection systems by utilizing readily available weakly supervised detection data. Although our method is built upon the specific application of pulmonary nodule detection, the framework itself is fairly general and can be readily applied to other medical image deep learning applications to take advantage of weakly labeled data.

\section*{Acknowledgement}
We gratefully acknowledge the sharing of pulmonary CT imaging data from National Lung Screening Trial (NLST). We also thank NVIDIA for supporting this research. 

\bibliographystyle{splncs03}
\small{\bibliography{typeinst}}
\end{document}